\PassOptionsToPackage{table}{xcolor}
\documentclass[conference,a4paper]{IEEEtran} 
\IEEEoverridecommandlockouts
\usepackage{cite}
\usepackage{comment}
\usepackage{enumitem}
\usepackage{amsmath,amsfonts,amssymb}
\usepackage[ruled,vlined,linesnumbered]{algorithm2e}
\usepackage{bm}
\usepackage{booktabs}
\usepackage{multirow}
\usepackage{graphicx}
\usepackage{subfig}
\usepackage{scrextend} 
\usepackage[usestackEOL]{stackengine}
\usepackage{url} 
\usepackage{array}
\usepackage{makecell}
\usepackage{flushend}
\usepackage[normalem]{ulem}
\usepackage{tcolorbox}
\usepackage[table]{xcolor}

\newcommand{\PreserveBackslash}[1]{\let\temp=\\#1\let\\=\temp}
\newcolumntype{C}[1]{>{\PreserveBackslash\centering}p{#1}}
\newcolumntype{R}[1]{>{\PreserveBackslash\raggedleft}p{#1}}
\newcolumntype{L}[1]{>{\PreserveBackslash\raggedright}p{#1}}

\SetKwIF{If}{ElseIf}{Else}{if}{}{else if}{else}{end if}%

\definecolor{darkspringgreen}{rgb}{0.09, 0.45, 0.27}
\definecolor{darkred}{rgb}{0.55, 0.0, 0.0}
\definecolor{darkgoldenrod}{rgb}{0.72, 0.53, 0.04}
\definecolor{denim}{rgb}{0.08, 0.38, 0.74}

\begin{document}

\title{AiGAS-dEVL-RC: An Adaptive Growing Neural Gas Model for Recurrently Drifting\\Unsupervised Data Streams}

\author{\IEEEauthorblockN{Maria Arostegi\IEEEauthorrefmark{1}, Miren Nekane Bilbao\IEEEauthorrefmark{2}, Jesus L. Lobo\IEEEauthorrefmark{1}, and Javier Del Ser\IEEEauthorrefmark{1}\IEEEauthorrefmark{2}}
\IEEEauthorblockA{\IEEEauthorblockA{\IEEEauthorrefmark{1}TECNALIA, Basque Research and Technology Alliance (BRTA), 48160 Derio, Bizkaia, Spain\\}
\IEEEauthorblockA{\IEEEauthorrefmark{2}University of the Basque Country (UPV/EHU), 48940 Leioa, Bizkaia, Spain\\}
Email: \{maria.arostegi, jesus.lopez, javier.delser\}@tecnalia.com, nekane.bilbao@ehu.eus
}
}

\IEEEoverridecommandlockouts
\IEEEpubid{\makebox[\columnwidth]{\copyright2025 IEEE. Personal use of this material is permitted.\hfill} \hspace{\columnsep}\makebox[\columnwidth]{ }}

\maketitle

\IEEEpubidadjcol

\begin{abstract}
Concept drift and extreme verification latency pose significant challenges in data stream learning, particularly when dealing with recurring concept changes in dynamic environments. This work introduces a novel method based on the Growing Neural Gas (GNG) algorithm, designed to effectively handle abrupt recurrent drifts while adapting to incrementally evolving data distributions (incremental drifts). Leveraging the self-organizing and topological adaptability of GNG, the proposed approach maintains a compact yet informative memory structure, allowing it to efficiently store and retrieve knowledge of past or recurring concepts, even under conditions of delayed or sparse stream supervision. Our experiments highlight the superiority of our approach over existing data stream learning methods designed to cope with incremental non-stationarities and verification latency, demonstrating its ability to quickly adapt to new drifts, robustly manage recurring patterns, and maintain high predictive accuracy with a minimal memory footprint. Unlike other techniques that fail to leverage recurring knowledge, our proposed approach is proven to be a robust and efficient online learning solution for unsupervised drifting data flows.
\end{abstract}

\begin{IEEEkeywords}
Data stream learning, extreme verification latency, concept drift, Growing Neural Gas.
\end{IEEEkeywords}

\section{Introduction} \label{sec:intro}

Data stream learning has become increasingly relevant in a variety of real-world applications, ranging from fraud detection and stock market analysis to personalized recommendations and industrial process monitoring \cite{vzliobaite2016overview}. These systems rely on continuous real-time processing of data streams to make predictions or decisions. Unlike static datasets, data streams are often characterized by their unbounded, high-speed nature, which necessitates models that can operate incrementally, efficiently, and with minimal reliance on labeled data. Ensuring that such models remain accurate and adaptive over time is crucial for maintaining the performance of systems operating in dynamic environments \cite{gama2012survey,zliobaite2012next,gomes2019machine}.

In this research area, Extreme Verification Latency (EVL) refers to streaming scenarios where ground-truth labels for data points arrive with significant delays or may be completely unavailable for extended periods \cite{marrs2010impact}. This phenomenon is common in domains such as medical diagnostics, where true outcomes may take weeks or months to materialize, or in cybersecurity, where labeling attacks requires detailed forensic analysis. EVL is challenging for data stream learning systems, as delayed feedback makes it difficult to update models promptly, leading to potential degradation in performance. Designing methods capable of learning effectively under such constraints is vital for ensuring the reliability of predictive models in these contexts \cite{souza2015data}.

Another challenging phenomenon in data stream learning is Concept Drift (CD), which occurs when the underlying data distribution changes over time, rendering previously learned models inaccurate \cite{gama2014survey,ditzler2015learning}. CD can be incremental, gradual, sudden, or recurring, and arises in virtually all dynamic environments, from customer behavior analysis to environmental monitoring. Adapting to CD is a well-studied problem in data stream learning, with various methods developed to detect, accommodate, or mitigate its effects \cite{lu2018learning,read2022learning}. However, managing drift becomes increasingly complex when coupled with EVL, as the delayed arrival of labels hinders the model's ability to promptly adjust to distributional changes.

While there has been some research addressing setups that involve both CD and EVL, these studies remain limited in scope. In particular, they often fail to account for scenarios where the drift is recurrent. Recurrent CD refers to situations where previously encountered data distributions reappear after a period of absence \cite{gama2014recurrent}. Assuming a data stream $\{(\mathbf{x}_t, y_t)\}_{t=1}^\infty$ with \(P_t(\mathbf{x}, y)\) denoting the joint probability distribution at time \(t\), an abrupt recurrent CD can be defined by a sequence of time indices \(t_1, t_2, \ldots, t_k\) where abrupt drifts occur, such that for any two distinct drift points \(t_i\) and \(t_j\) (\(i \neq j\)): (1) $P_{t_i}(\mathbf{x}, y) \neq P_{t_i-1}(\mathbf{x}, y)$; (2) $P_{t_j}(\mathbf{x}, y) \neq P_{t_j-1}(\mathbf{x}, y)$; and (3) for some \(t_i, t_j\), $P_{t_i}(\mathbf{x}, y) = P_{t_j}(\mathbf{x}, y)$. Such patterns are common in seasonal data (e.g., retail sales trends or environmental data) and cyclic processes (e.g., production cycles in manufacturing). The ability to efficiently store and retrieve knowledge of these recurring patterns is crucial for building robust systems \cite{gunasekara2024recurrent}.

Adapting to abrupt recurrent CD under EVL poses unique challenges. The absence of supervisory feedback hinders the timely recognition of recurring patterns, as models lack immediate validation to characterize the prevalent distribution $P_t(\mathbf{x},y)$ and confirm the reappearance of a prior concept. Additionally, efficiently managing memory to balance the retention of historical knowledge with responsiveness to new drifts is complex, particularly when the system operates under memory and/or processing constraints. Without suitable mechanisms to identify and leverage past knowledge, models struggle when relearning recurring patterns, leading to inefficiencies in computational resources and lower predictive accuracy. 

Despite its practical importance, the combined challenge of recurrent CD and EVL has been largely overlooked in the literature, leaving a critical gap that this paper seeks to address. Specifically, we propose a novel methodology based on Growing Neural Gas (GNG) for streaming setups subject to EVL and CD, including incremental and abrupt recurrent drifts. GNG, an incremental artificial neural network that learns topological relationships, is particularly well-suited to handle EVL and incremental CD by dynamically adapting its structure to evolving data distributions. To tackle abrupt and recurrent CD, our methodology -- hereafter coined as AiGAS-dEVL-RC -- incorporates a mechanism for storing representative nodes produced by GNG, enabling the system to efficiently retrieve and reuse previously learned knowledge when recurring patterns are detected. The proposed approach is evaluated on datasets designed to simulate incremental and abrupt CD, allowing the model to adapt incrementally to dynamic situations and demonstrating its effectiveness in better managing abrupt and recurring changes when compared to other EVL approaches for drifting data streams.

The rest of the paper is structured as follows: Section \ref{sec:related_work} first reviews the state of the art, with a focus on EVL and recurrent CD. Next, a detailed explanation of the proposed AiGAS-dEVL-RC method is given in Section \ref{sec:approach}. The experimental setup and the datasets in use are specified in Section \ref{sec:experiments}, while the results of our experiments are presented and discussed in Section \ref{sec:results}. Section \ref{sec:conclusions} concludes the article with a summary of our findings and directions for future research.

\section{Related Work and Contribution} \label{sec:related_work}

Before proceeding with the description of the proposed approach, we define and contextualize the key concepts central to this research, including EVL and CD (Subsection \ref{ssec:evl_cd}), recurrent CD (Subsection \ref{ssec:rcd}) and knowledge storage/retrieval under such circumstance (Subsection \ref{ssec:ksr}), and GNG (Subsection \ref{ssec:gng}). We end our literature review on these central topics for our research with a short statement of the novelty of this work within the reviewed literature (Subsection \ref{ssec:contrib}).

\subsection{Extreme Verification Latency and Concept Drift} \label{ssec:evl_cd}

CD occurs when the underlying data distribution $P_t(\mathbf{x},y)$ evolves over time, altering the relationships between input features and target variables. This phenomenon can manifest in various forms, including sudden (sharp) drifts caused by abrupt, unforeseen events, or incremental drifts that evolve progressively over time. Each type of drift induces unique challenges into the design of online (machine) learning models, necessitating continuous and adaptive learning strategies to maintain their predictive performance. Adaptation approaches to CD can be broadly categorized into \emph{active} and \emph{passive} methods. Active strategies detect drift explicitly, triggering retraining or updates, while passive strategies incrementally adapt models to the evolving data distribution without explicit detection mechanisms. Both approaches aim to ensure model robustness in the dynamic nature of streaming data environments \cite{zliobaite2012next,krempl2014open,ditzler2015learning,lu2018learning}.

In data stream learning, EVL arises in streaming scenarios where labeled data arrives sparsely or with significant delays, if at all. This absence of immediate feedback hinders model updates, increasing the risk of performance degradation as data distributions evolve. Adaptation mechanisms under EVL often leverage semi-supervised or active learning paradigms to mitigate this challenge. Semi-supervised approaches, such as COMPOSE \cite{dyer2013compose} and AMANDA \cite{ferreira2019amanda}, use limited labeled samples alongside abundant unlabeled data to infer patterns, while active learning selectively queries instances to optimize learning efficiency. EVL demands strategies that maximize utility from scarce supervision, enabling models to adapt despite delayed or missing labels \cite{siekmann2007knowledge,marrs2010impact,umer2017learning}.

Recent advancements tackle both incremental CD and EVL simultaneously, focusing on solutions for non-stationary data streams with minimal supervision. In this context we highlight \cite{umer2020comparative}, which provides a review and comparison of techniques proposed to deal with such streaming scenarios. Among them, techniques like SCARGC \cite{souza2015data} and FAST COMPOSE \cite{umer2016learning} enhance clustering-based drift adaptation while optimizing computational efficiency. LEVELIW \cite{umer2017level} introduces importance weighting for iterative updates, ensuring alignment with evolving concepts. AMANDA extends density-based clustering to dynamically adapt to distribution shifts, and TRACE \cite{arostegi2018concept} leverages trajectory prediction for tracking incremental drifts. Building on TRACE, SLAYER \cite{arostegi2021slayer} incorporates incremental clustering and distance-based matching to better handle dynamic streams with variable concept distributions. 

\subsection{Recurrent Concept Drift} \label{ssec:rcd}

There are numerous examples in the real world where, due to the seasonality of the features involved in the definition of the dataset (e.g., those related to weather events, financial markets, sensor data from IoT devices, or customer behavior in e-commerce), patterns evolve and reoccur over time. As defined in the introduction, recurrent CD refers to a situation in machine learning (ML) where the underlying relationship between input features $\mathbf{x}$ and the target variable $y$ changes over time. Unlike one-time drifts, recurrent CDs indicate that the system revisits earlier states or patterns in a cyclic or repetitive manner. Consequently, the system must adapt not only to the current drift, but also to the possibility that past data distributions may reappear. This type of drift can be particularly challenging for ML models because it requires ongoing adaptation to fluctuating data patterns without assuming a one-time shift. The model must learn from past drifts and adapt to the recurrent nature of the changes.

The design of efficient strategies for knowledge persistence and retrieval in data stream learning has grasped the interest of the scientific community over the years. This is evident in the early work by Gomes et al. \cite{gomes2013mining}, who emphasize the importance of efficiently using a memory of stored models, and propose a scheme to adapt to recurrence by associating context with target concepts. Regardless of the model used, one of the most widely adopted approaches to identify past concepts from stream data is the selection of the most relevant features, the correlation between observed and stored features, or the use of \emph{meta-features} \cite{gomes2013mining,hu2019algorithm,ahmadi2018modeling,halstead2023combining}. Another approach is to focus on \emph{concepts} rather than raw data instances, so that associating them with internal states helps to evaluate the relevance of past experience \cite{sripirakas2014mining, halstead2022probabilistic}. Another work worth mentioning is \cite{zheng2021semi}, which presents a semi-supervised learning framework that takes into account the evolution of concepts by monitoring outliers and analyzing their cohesion and is capable of detecting new classes. When it comes to extracting useful information from the stream and updating it with persisted knowledge, most of the literature resorts to ensemble methods composed of dynamically weighted learners \cite{museba2021recurrent}, meta-models \cite{angel2016predicting}, and meta-learning techniques \cite{gama2014recurrent}. Finally, we pause at the work in \cite{gunasekara2024recurrent}, which delves into ML definitions for data streams affected by recurrent conceptual drifts, and methodological approaches clarifying its key design components. In addition, it explores various evaluation techniques, benchmark datasets, and available software adapted to reproduce and analyze data streams with recurrent CDs. Further insights into this profitable area of research can be found in the recent survey in \cite{suarez2023survey}, which provides a complete description of the detection and adaptation mechanisms used to deal with recurrent CDs, as well as a list of models to deal with this particular flavor of non-stationary data streams, including online ensembles, meta-learning, and model-based clustering. 

\subsection{Knowledge Storage/Retrieval for Recurrent Concept Drift} \label{ssec:ksr}

A key advantage of designing mechanisms to recognize or remember prior situations (\emph{knowledge}) is the ability to reuse ML models without retraining, leading to significant computational savings, particularly in fully unsupervised scenarios where starting from scratch is infeasible. However, efficiently storing these models for future recognition is critical to avoid memory-related issues.

Early work in this area \cite{anderson2016cpf} introduced a meta-learner that employs a concept drift detector alongside a pool of models. Upon detecting drift, the stored models are evaluated based on their accuracy. Depending on this evaluation, a model is either reused or replaced by a newly trained one, which is then stored for future use. While this approach addresses storage and reuse, the selection of appropriate classifiers remains a challenge. To tackle this, \cite{anderson2019recurring} proposed the so-called \emph{Enhanced Concept Profiling Framework} (ECPF), which identifies recurring concepts and reuses previous classifiers based on classification similarity with incoming data. This framework reduces the time required to select suitable classifiers, offering a more efficient solution. Another proposal in this area is \cite{he2020incremental}, which proposes an incremental learning framework generating a sequence of models capable of mitigating the CD by updating all classes with new instances.

Further advancements, as discussed in \cite{halstead2021recurring}, reflect on policies for deciding whether a model should be stored. Three key factors are proposed to guide this decision: (i) the frequency of future recurrences, (ii) the advantage of retaining the model, and (iii) transparency in storage policies. Building on this, \cite{halstead2023combining} introduces a framework that integrates diverse meta-features into a single representation. This method not only enhances computational and storage efficiency, but also dynamically identifies which meta-features best distinguish concepts in a given dataset, significantly improving overall performance.

In this work, we address the challenges posed by the absence of supervision (EVL) and abrupt recurrent CDs. When prior knowledge cannot be leveraged, it is crucial to characterize concepts over time in an unsupervised manner and design algorithmic criteria to retrieve them when they emerge again from the stream. This involves balancing the accuracy of profiling recurring patterns with the memory required to store them for subsequent retrieval.

\subsection{Growing Neural Gas} \label{ssec:gng}

GNG \cite{sun2017online} is an incrementally learnable neural network that characterizes topological relations in data using competitive Hebbian learning. Unlike \emph{Self-Organizing Maps} (SOMs), which use a predefined structure, GNG defines neighborhoods based on distances in the input space, allowing for more flexible adaptation. A key feature of GNG is its incremental learning capability, which enables continuous adaptation to new data, making it particularly suitable for processing data streams. Furthermore, GNG does not require a number of neurons to be established beforehand; instead, it adds new units as needed to discover the optimal network structure. This approach allows the algorithm to create a graph that can be visually represented, revealing underlying cluster patterns in complex multidimensional datasets.

Decades after its inception \cite{fritzke1994growing}, recent research has highlighted the potential of GNGs across multiple domains, including exploratory data analysis and multidimensional data scaling \cite{ventocilla2021scaling}, image and video processing \cite{dale2024streamsongv2}, planning in autonomous robotics \cite{Ozasa:24}, and adaptation to incremental CD in dynamic environments \cite{bouguelia2018adaptive,arostegi2018concept}. The algorithm's most compelling attribute for unsupervised learning from data streams lies in its ability to dynamically update its topological structure. As new data arrives, GNG can continuously modify its neural network architecture, creating a responsive and adaptive model that captures evolving patterns in the input space with remarkable flexibility and precision. Remarkably, several improved versions of GNG have appeared in recent literature, including its learning speed and convergence when capturing patterns at different scales \cite{obo2024fast}.

\subsection{Contribution} \label{ssec:contrib}

Recent research \cite{arostegi2024aigas} underscores the capability of GNG to effectively characterize streaming data, even in the absence of labels, while adapting robustly to incremental concept drifts. Building on this foundation, we propose that GNG can identify singular nodes that capture the feature space of dominant concepts. By storing these nodes alongside additional information, such as cluster centroid positions, and retrieving them when needed, instance-based learners can better address recurrent drifts in unsupervised data streams. This work introduces a novel approach for leveraging these singular nodes and their associated cluster data to enhance model retention and recurrence detection. Specifically, we utilize the Intersection over Union (IoU) metric to assess whether the current concepts in the data stream align with those previously characterized by GNG. When alignment is detected, the corresponding stored distribution of concepts is retrieved to facilitate accurate predictions for incoming data. This approach mitigates model divergence and supports adaptive learning in dynamic environments, including those subject to EVL, incremental CD, and abrupt recurrent CD.

\section{Proposed AiGAS-dEVL-RC Algorithm}\label{sec:approach}

As explained in Section \ref{sec:related_work}, the proposed algorithm is based on the use of GNG as a method to characterize the stream instances over time, looking for the feature representation space that best defines the shapes and distributions of all the concepts detected in the stream. AiGAS-dEVL-RC builds upon this observation, and incorporates procedures for storing and retrieving prior knowledge, alongside the criterion to detect that a recurrent CD occurs in the stream. 
\vspace{-2mm}
\begin{figure}[h!]
    \centering
\includegraphics[width=\columnwidth]{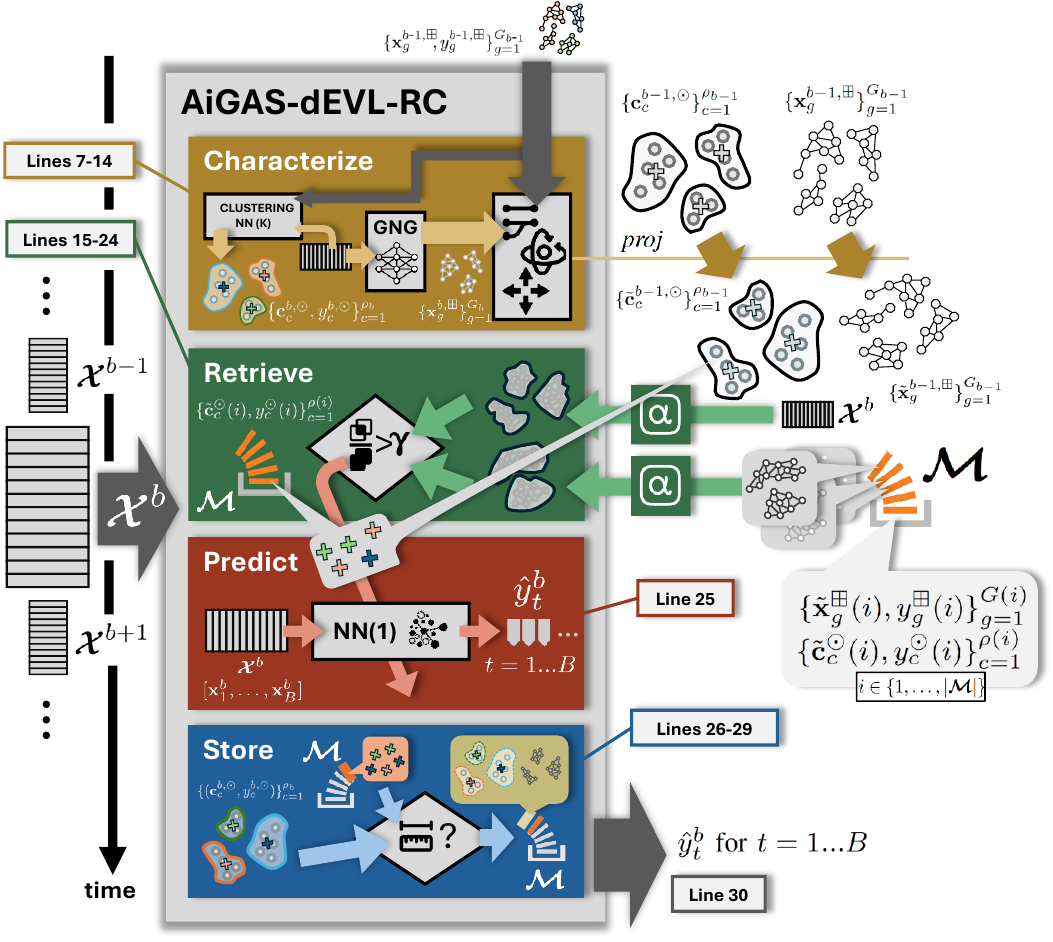}
    \caption{Diagram showing the main steps of AiGAS-dEVL-RC.}
    \label{fig:diagram}
\end{figure}

Figure \ref{fig:diagram} depicts a graphical summary of the proposed approach, which is further complemented by the detailed algorithmic description in Algorithm \ref{alg:ai_gas_devl}. The design of AiGAS-dEVL-RC hinges on 4 different processing stages applied on every single batch of streaming instances received over time: 1) {\color{darkgoldenrod}\textbf{characterize}}, which aims to characterize the distribution of concepts in the current batch and estimate (\emph{project}) how instances drift incrementally over time; 2) {\color{darkspringgreen}\textbf{retrieve}}, which examines a memory of stored concepts and identifies whether any distribution in the past overlaps with the distribution of concepts within the current batch; 3) {\color{darkred}\textbf{predict}}, which elicits the labels predicted for the current batch instances based on the projected or the retrieved distribution of concepts; and 4) {\color{denim}\textbf{store}}, where the algorithm decides whether the distribution of concepts in the current batch is novel enough for its storage in the memory. Colors of the steps in the figure are matched to those of the comments in the algorithmic description, and lines are also indicated in the figure so as to denote which processing operations are involved in each stage. 

From a general perspective, AiGAS-dEVL-RC relies on the use of an instance-based model to {\color{darkred}\textbf{predict}} the instances within a newly arriving batch in the stream, which implies that the \emph{knowledge} captured by the model is embedded in a set of data instances retained by the model over time. This strategy goes in line with the importance of instance-based adaptation mechanisms for data streams highlighted in recent surveys \cite{suarez2023survey}. Such stored samples are used to predict new stream samples using a prototype-based model, by which stream instances are compared to a representative set of samples, which summarize the distribution of concepts within each class (e.g., through centroids, medoids, or more advanced representations). 

Given the EVL and incremental drift assumptions, it is necessary to ensure that the algorithm can estimate the trajectory delineated by the different concepts (and their supervision) over time. This motivates the previous {\color{darkgoldenrod}\textbf{characterization}} phase of AiGAS-dEVL-RC, which resorts to the projection-based approach of its predecessor AiGAS-dEVL \cite{arostegi2024aigas}. Assuming stream instances are collected in batches of finite size $\bm{\mathcal{X}}^b=\{\mathbf{x}_t^b\}_{t=1}^B$ (with $b=1,\ldots,\infty$), GNG nodes $\{\mathbf{x}_g^{b,\boxplus}\}_{g=1}^{G_b}$ are first extracted from the instances within the batch, which are then used to compute a \emph{projection}\footnote{As discussed in \cite{arostegi2024aigas}, different projections can be learned from the GNG nodes of the batches depending on the incremental drift present in the stream.} that relates them to the GNG nodes of the previous batch $\{\mathbf{x}_g^{b-1,\boxplus}\}_{g=1}^{G_{b-1}}$. Then, GNG nodes of the current batch and the prototypes are projected based on the learned transformation, serving as an estimation of the evolution of the drifted concepts that exist in the stream over time. By applying the projection also on prototypes, predictions of the next batch account for the dynamics of the drift, endowing it with the capability to accommodate incremental drifts in an unsupervised fashion.

The assumption that drifts can be also abrupt and recurrent requires further algorithmic modifications and extensions. In this regard, AiGAS-dEVL-RC extends the above projection-based mechanism by equipping the algorithm with two phases, {\color{darkspringgreen}\textbf{retrieve}} and {\color{denim}\textbf{store}}, which are based on a memory $\bm{\mathcal{M}}$ of past models. In the context of instance-based prediction, a \emph{model} refers to all the information necessary to (i) predict new samples (prototypes and their estimated classes) and (ii) characterize the shape and distribution of predicted classes across different concepts within the data stream (GNG nodes and their estimated labels). A key design element of AiGAS-dEVL-RC is the criterion for comparing and recognizing the similarity between the current concept distribution and previously stored concept distributions. To this end, much of the existing literature relies on the statistical analysis of distributions, including statistical multivariate likelihood tests \cite{ahmadi2018modeling} or non-parametric multivariate statistical tests \cite{gonccalves2013rcd}. Differently, AiGAS-dEVL-RC employs a threshold based on the Intersection over Union (IoU) metric, computed between the $\alpha$-shapes of the stored GNG nodes and the stream instances of the current batch:
\begin{itemize}[leftmargin=*]
\item On one hand, $\alpha$-shapes are a subset of the Delaunay triangulation that captures the shape of a point set by filtering simplices based on a parameter \(\alpha\). Simplices are retained if their circumscribing spheres have a radius smaller than or equal to \(\alpha\). As \(\alpha\) varies, the alpha-shape transitions from a detailed representation to the convex hull of the point set. $\alpha$-shapes are widely used in science and engineering applications, including structural molecular biology \cite{gardiner2018alpha} or the volumetric characterization of tumors \cite{al2015alpha}. Recent research aims to enhance their computational efficiency \cite{edelsbrunner2023simple}.

\item On the other hand, the IoU metric measures the overlap between two arbitrary shapes, and is widely used to measure the performance of any object detection method by comparing the ground truth bounding box to the one predicted by the object detection model. This metric is computed as the ratio of the area of overlap between two shapes (e.g., a predicted bounding box and its ground truth) to the area of their union, providing a measure of how accurately they align with each other.
\end{itemize} 

By computing $\alpha$-shapes of the GNG nodes within the current stream and comparing them to those in the memory of past models (using the IoU metric), AiGAS-dEVL-RC can detect whether the drift has evolved into a distribution of concepts in the stream that resembles one already encountered in the past. When an abrupt recurrent drift occurs, causing the stream to revert suddenly to a past concept distribution, AiGAS-dEVL-RC retrieves information from its memory of stored models. This enables the algorithm to maintain robust performance, even when the stream undergoes sudden changes.
\begin{algorithm}[!h]
	\caption{Proposed AiGAS-dEVL-RC algorithm.}\label{alg:ai_gas_devl}
	\footnotesize
	\DontPrintSemicolon
	\SetAlgoLined
	\SetKwInOut{Input}{Input}
	\SetKwInOut{Output}{Output}	
    \Input{Initially supervised ($\{\mathbf{x}_t,y_t\}_{t<T_s}$) and   
    unsupervised ($\{\mathbf{x}_t\}_{t\geq T_s}$) stream instances, batch size $B$, $\Delta$ (used to control the models to be stored), $\gamma$ ($\alpha$-shape similarity threshold), $\epsilon$ (maximum distance between centroids), $\bm{\mathcal{M}}=\emptyset$ (memory), $\epsilon_\Delta$, $\epsilon_D$, $\gamma$.}	
	\Output{Predicted labels $\{\widehat{y}_t\}_{t\geq T_s}$.}
    \vspace{1mm}
    \tcp{Initially supervised part of the stream}
    \vspace{1mm}
    Compute prototypes $\{(\mathbf{c}_c^{0,\odot},y_c^{0,\odot})\}_{c=1}^{\rho_0}$ from $\{\mathbf{x}_t,y_t\}_{t<T_s}$\;
    Compute GNG nodes $\{\mathbf{x}_g^{0,\boxplus}\}_{g=1}^{G_0}$ from $\{\mathbf{x}_t\}_{t<T_s}$\;
    Predict GNG node labels: $y_g^{0,\boxplus}=\textup{NN}(\mathbf{x}_g^{0,\boxplus};\{\mathbf{x}_t,y_t\}_{t<T_s},K)$\;
    Initialize: $\tilde{\mathbf{x}}_g^{0,\boxplus}=\mathbf{x}_g^{0,\boxplus}$ ($g=1...G_0$), $\tilde{\mathbf{c}}_c^{0,\odot}=\mathbf{c}_c^{0,\odot}$ ($c=1...\rho_0$)\;
    Store: $\bm{\mathcal{M}} \leftarrow [\{(\tilde{\mathbf{c}}_c^{0,\odot},\mathbf{y}_c^{0,\odot})\}_{c=1}^{\rho_0},\{(\tilde{\mathbf{x}}_g^{0,\boxplus},y_g^{0,\boxplus})\}_{g=1}^{G_0}]$\;
    \vspace{1mm}
    \tcp{Unsupervised stream batches}
    \vspace{1mm}
	\For{$b\in[1,\ldots,\infty)$} 
	{
	{\color{darkgoldenrod}\tcp{\textbf{Characterize}}}
        Collect a new batch as $\bm{\mathcal{X}}^b\doteq[\mathbf{x}_1^b,\ldots,\mathbf{x}_B^b]$\;
        Compute GNG nodes $\{\mathbf{x}_g^{b,\boxplus}\}_{g=1}^{G_b}$ from $\bm{\mathcal{X}}^b$\;
        Annotate GNG nodes as: $y_g^{b,\boxplus}=\textup{NN}(\mathbf{x}_g^{b,\boxplus};\{\widetilde{\mathbf{x}}_g^{b-1,\boxplus},y_g^{b-1,\boxplus}\}_{g=1}^{G_{b-1}},K)$ ($g=1...G_b$)\;
        Compute prototypes $\{(\mathbf{c}_c^{b,\odot},y_c^{b,\odot})\}_{c=1}^{\rho_b}$ from $\bm{\mathcal{X}}^b$\;   
        Annotate prototypes as: $y_c^{b,\odot}=\textup{NN}(\mathbf{c}_c^{b,\boxplus};\{\widetilde{\mathbf{x}}_g^{b-1,\boxplus},y_g^{b-1,\boxplus}\}_{g=1}^{G_{b-1}},K)$ ($c=1...\rho_b$)\;        
        Estimate \emph{proj} from $\{\mathbf{x}_g^{b,\boxplus}\hspace{-1mm},y_g^{b,\boxplus}\}_{g=1}^{G_b}$, $\{\mathbf{x}_g^{b\mbox{-}1,\boxplus}\hspace{-1mm},y_g^{b\mbox{-}1,\boxplus}\}_{g=1}^{G_{b\mbox{-}1}}$ \cite{arostegi2024aigas}\;
        Project previous centroids: $\{\mathbf{c}_c^{b-1,\odot}\}_{c=1}^{\rho_{b-1}}\xrightarrow{\textup{proj}} \{\tilde{\mathbf{c}}_c^{b-1,\odot}\}_{c=1}^{\rho_{b-1}}$\;
        Project previous nodes: $\{\mathbf{x}_g^{b-1,\boxplus}\}_{g=1}^{G_{b-1}}\xrightarrow{\textup{proj}}  \{\tilde{\mathbf{x}}_g^{b-1,\boxplus}\}_{g=1}^{G_{b-1}}$\;
        {\color{darkspringgreen}\tcp{\textbf{Retrieve}}}
        \For{$i\in\{1,\ldots,|\bm{\mathcal{M}}|\}$}
            {
            \If{$\nexists  c\hspace{-.5mm}\in\hspace{-.5mm}\{1...\rho_b\}: \min_{k\{1...|\bm{\mathcal{M}}(i)|\}}\hspace{-.5mm}D(\mathbf{c}_c^{b,\odot}\hspace{-.5mm},\tilde{\mathbf{c}}_{k}^{\odot}(i))\hspace{-.5mm}>\epsilon_R$}
            {
            Compute $\alpha$-shapes of $\bm{\mathcal{X}}^b$ and $\{\tilde{\mathbf{x}}_g^{\boxplus}(i)\}_{g=1}^{G(i)}$\;
            \If{$\textup{IoU between such $\alpha$-shapes}>\gamma$}
            {
                
                $\{\tilde{\mathbf{x}}_g^{b-1,\boxplus},y_g^{b-1,\boxplus}\}_{g=1}^{G_{b-1}}\leftarrow \{\tilde{\mathbf{x}}_g^{\boxplus}(i),y_g^{\boxplus}(i)\}_{g=1}^{G(i)}$\;
                $\{\tilde{\mathbf{c}}_c^{b-1,\odot},y_c^{b-1,\odot}\}_{c=1}^{\rho_{b-1}}\leftarrow\{\tilde{\mathbf{c}}_c^{\odot}(i),y_c^{\odot}(i)\}_{c=1}^{\rho(i)}$\;
                \textbf{break}\;
            }
            }
            }
        {\color{darkred}\tcp{\textbf{Predict}}}
        $\widehat{y}_t^b=\textup{NN}(\mathbf{x}_t^b;\{\tilde{\mathbf{c}}_c^{b-1,\odot},y_c^{b-1,\odot}\}_{c=1}^{\rho_{b-1S}},1)$ ($t=1...B$)\; 
        {\color{denim}\tcp{\textbf{Store}}}
        Let $i^\ast=|\bm{\mathcal{M}}|$ (index of last item in memory $\bm{\mathcal{M}}$)\;
\If{$\exists c\hspace{-.3mm}\in\hspace{-.3mm} \{1...\rho_b\}: \min_{k\in\{1...|\bm{\mathcal{M}}(i^\ast)|\}}D(\mathbf{c}_c^{b,\odot},\tilde{\mathbf{c}}_{k}^{\odot}(i^\ast)\hspace{-.3mm}>\hspace{-.3mm}\epsilon_D$}
        {
            Store: $\bm{\mathcal{M}} \leftarrow [\{(\mathbf{c}_c^{b,\odot},\mathbf{y}_c^{b,\odot})\}_{c=1}^{\rho_b},\{(\mathbf{x}_g^{b,\boxplus},y_g^{b,\boxplus})\}_{g=1}^{G_b}]$\;
        }
        Return $\hat{y}_t^b$ for $t=1...B$ (line \textbf{24}), and proceed with batch $b+1$\;
        }        
\end{algorithm}

Algorithm \ref{alg:ai_gas_devl} summarizes the main steps of the proposed AiGAS-dEVL-RC algorithm. First, prototypes $\{\tilde{\mathbf{c}}_c^{0,\odot},y_c^{0,\odot}\}_{c=1}^{\rho_{0}}$ are extracted from the initially supervised data instances $\{(\mathbf{x}_t,y_t)\}_{t<T_s}$ (line \textbf{1}) by using a clustering algorithm. Labels are computed based on the composition of the resulting clusters. Likewise, GNG nodes $\{\mathbf{x}_g^{0,\boxplus}\}_{g=1}^{G_0}$ are extracted over from this initially supervised set of instances (line \textbf{2}) and labeled (line \textbf{3}) using a $K$ nearest neighbors classifier $\textup{NN}(a;b,K)$ (with $a$ denoting the query instance, $b$ the reference dataset, and $K$ the number of neighbors), using the initially supervised instances as the reference dataset. After initializing prototypes and GNG nodes (line \textbf{4}) and storing them in the memory $\bm{\mathcal{M}}$ (line \textbf{5}), AiGAS-dEVL-RC iterates on every single batch $\bm{\mathcal{X}}^b$ received from the stream by following sequentially the four phases described previously: 
\begin{itemize}[leftmargin=*]
\item {\color{darkgoldenrod}\textbf{\texttt{Characterize}}}: GNG nodes $\{\mathbf{x}_g^{b,\boxplus}\}_{g=1}^{G_b}$ and prototypes $\{\mathbf{c}_c^{b,\odot}\}_{c=1}^{\rho_b}$ are computed (lines \textbf{8} and \textbf{10}) over the batch instances using a clustering algorithm and GNG, respectively, and annotated (lines \textbf{9} and \textbf{11}) using the projected GNG nodes $\{\mathbf{x}_g^{b-1,\boxplus}\}_{g=1}^{G_{b-1}}$ from the previous batch. Then, a projection is estimated as in \cite{arostegi2024aigas} (line \textbf{12}) and applied to both centroids (line \textbf{13}) and GNG nodes (line \textbf{14}) of the previous batch, so that they better anticipate the drift dynamics incrementally evolving in the stream. The projected prototypes are fed to the {\color{darkred}\textbf{\texttt{predict}}} phase as the reference dataset used to predict the instances within the current batch.

\item {\color{darkspringgreen}\textbf{\texttt{Retrieve}}}: once the current batch has been characterized, AiGAS-dEVL-RC searches for past concept distributions in the stream that \emph{resemble} the prevalent one. To this end, the algorithm implements a two-step criterion: first, a threshold $\epsilon_R$ imposed on a measure of distance $D(\cdot,\cdot)$ between the centroids of the current batch and those stored in each of the distributions inside the memory (line \textbf{16}), and second, a threshold $\gamma$ on the IoU between the $\alpha$-shapes computed over the stream instances and the GNG nodes of each distribution in $\bm{\mathcal{M}}$ (lines \textbf{17} and \textbf{18}). Such distances are computed over all pairs of centroids, such that if all distance values fall below their corresponding threshold, a match is declared between the current distribution of concepts and a past one stored in the memory. In that case, the projected prototypes and GNG nodes are rewritten with the information stored in the memory (lines \textbf{19} and \textbf{20}). This allows AiGAS-dEVL-RC to update its reference dataset to a previous state into which the stream flowed incrementally, better accommodating sudden recurrent drifts in the absence of supervision.

\item {\color{darkred}\textbf{\texttt{Predict}}}: in this third phase, the labels for the stream instances in the present batch are predicted using a reference dataset composed by prototypes. Such prototypes can be the projected prototypes of the previous batch (line \textbf{13}) or, alternatively, prototypes retrieved from the memory $\bm{\mathcal{M}}$ corresponding to a previous distribution of concepts similar to that characterized from the current batch (line \textbf{20}). Given its instance-based nature, and without loss of generality, AiGAS-dEVL-RC utilizes a nearest neighbor classifier with $K=1$ using the projected/retrieved centroids to elicit its predictions. 

\item {\color{denim}\textbf{\texttt{Store}}}: finally, this fourth phase decides whether the distribution of concepts modeled inside the current batch should be stored in the memory $\bm{\mathcal{M}}$. In doing so, AiGAS-dEVL-RC compares the last distribution saved in the memory with the one characterized from the current batch, using a distance threshold $\epsilon_D$ between pairs of prototypes conforming such distributions (line \textbf{27}). When the threshold is surpassed for any given pair of prototypes, the set of centroids and GNG nodes of the current batch (and their corresponding annotated labels) are stored in the memory $\bm{\mathcal{M}}$ (line \textbf{28}), becoming themselves the most recent distribution for the {\color{denim}\textbf{\texttt{store}}} phase of subsequent batches.
\end{itemize}

\section{Experimental Setup} \label{sec:experiments}

A set of experiments has been designed to assess the performance of the proposed AiGAS-dEVL-RC algorithm and to compare it to several methods from the state of the art in EVL and incremental CD in data streams revisited in Subsection \ref{ssec:evl_cd}. We consider the following baselines:
\begin{enumerate}[leftmargin=*]
\item A-FCP \cite{ferreira2019amanda}, which is a semi-supervised density-based adaptive model for non-stationary data, which selects a \textit{fixed} number of samples to be used as kernel to be used in the prediction of the next batch.
	
\item A-DCP \cite{ferreira2019amanda}, namely, an extension of A-FCP which considers a \textit{dynamic} number of kernel samples to predict the instances within new batches arriving from the stream.
	
\item AiGAS-dEVL \cite{arostegi2024aigas}, a recent semi-supervised adaptive modeling framework for non-stationary data streams that also hinges on GNG to characterize the shape and inner point distributions of all concepts detected within the stream over time. AiGAS-dEVL allows for the selection of the classifier type, matching algorithm, and node projection strategy, which can be tailored to the characteristics of the data stream and concept drift. However, AiGAS-dEVL proposed no methods for knowledge persistence and retrieval suited to deal with recurrent CD.

\item AiGAS-dEVL-RC, i.e., the approach proposed in this work, using K-means clustering as the method to extract prototypes (line \textbf{10} in Algorithm \ref{alg:ai_gas_devl}). Similarly to \cite{arostegi2024aigas}, both AiGAS-dEVL and AiGAS-dEVL-RC assume an Euclidean transformation \cite{wen2005total} by finding the optimal/best rotation and translation between the GNG nodes of consecutive batches (line \textbf{12} of the aforementioned algorithm). Table \ref{tab:datasets} shows the parameters' values of AiGAS-dEVL-RC used for every dataset; the rest of comparison baselines are configured as in the experiments reported in their respective publications.
\end{enumerate}
  \begin{table}[!h]
 	\centering
 	\caption{Description of the datasets used in the benchmark, including their number of classes and features, the total number of stream instances, batches and number of initially supervised instances ($t<T_s$). The last column indicates the values of the parameters defined in the algorithmic description of AiGAS-dEVL-RC (Algorithm \ref{alg:ai_gas_devl}.    \texttt{RCD} suffix indicates the modified version of the dataset with a recurrent CD induced at the end of the stream.}
 	\label{tab:datasets}
 	\resizebox{0.9\columnwidth}{!}{\begin{tabular}{cccc}
 		\toprule
 		\textbf{Dataset} & \makecell{\textbf{\# classes /}\\\# \textbf{features}} & \makecell{\textbf{\# of}\\\textbf{instances/batches/$T_s$}} & $\gamma$/$\epsilon_R$/$\epsilon_D$ \\
 		\midrule
 		\texttt{1CDT}   & 2/2 & 16,000/100/800 & 0.6/0.2/4.0\\
 		\texttt{1CHT}   & 2/2 &16,000/100/800 & 0.6/0.2/4.0\\
 		\texttt{2CDT}  & 2/2 & 16,000/100/800 & 0.6/0.2/4.0\\
 		\texttt{2CHT} & 2/2 & 16,000/100/800 & 0.6/0.2/4.0\\
 		\texttt{5CVT}   & 5/2 &24,000/200/1,000 & 0.6/0.2/1.5\\
 		\texttt{1CSURR}  & 2/2 &55,283/300/920 & 0.2/0.2/2.0\\
 		\texttt{MG2C2D} & 2/2 & 200,000/200/5,000 & 0.4/0.2/2.5\\
 		\texttt{FG2C2D} & 2/2  &200,000/200/5,000 & 0.6/0.2/1.5\\
 		\texttt{GEARS} & 2/2 &200,000/1,095/910 & 0.6/0.2/1.5\\
            \texttt{4CRT} & 4/2 & 144,400/100/7,220 & 0.6/0.1/1.5\\
            \texttt{4CRE-V1} & 4/2 & 125,000/500/1,250 & 0.7/0.1/4.0\\
            \texttt{4CRE-V2} & 4/2 & 183,000/800/1,140 & 0.7/0.1/4.0\\
            \texttt{UG2C2D} & 2/2 & 100,000/200/2,500 & 0.7/0.1/4.0\\
            \texttt{UG2C3D} & 2/3 & 200,000/200/2,000 & 0.7/0.1/4.0\\
            \texttt{UG2C5D} & 2/5 & 200,000/500/2,000 & 0.7/0.1/4.0\\
            \texttt{4CE1CF} & 5/2 & 173,250/200/4,330 & 0.7/0.1/4.0\\
            \midrule
            \texttt{1CDT-RCD} & 2/2 &20,000/100/1,000 & 0.6/0.2/4.0\\
 		\texttt{1CHT-RCD} & 2/2 &20,000/100/1,000 & 0.6/0.2/4.0\\
 		\texttt{2CDT-RCD} & 2/2 & 20,000/100/1,000 & 0.6/0.2/4.0\\
 		\texttt{2CHT-RCD} & 2/2 &20,000/100/1,000 & 0.6/0.2/4.0\\
 		\texttt{5CVT-RCD}  & 5/2 &30,000/200/750 & 0.3/0.2/1.5\\
 		\texttt{1CSURR-RCD}  & 2/2 & 60,000/300/1,000 & 0.2/0.3/2.0\\
 		\texttt{MG2C2D-RCD}  & 2/2 &218,900/200/5,500 & 0.4/0.2/2.5\\
 		\texttt{FG2C2D-RCD} & 2/2 &220,000/200/5,500 & 0.6/0.2/1.5\\
 		\texttt{GEARS-RCD}  & 2/2 &220,000/1,095/1,000 & 0.6/0.2/1.5\\
         \bottomrule
 	\end{tabular}}
 \end{table}
 
For the sake of fairness in the comparison between the above baselines, our benchmark must include stream datasets with incremental and abrupt recurring CDs. For this purpose, experiments consider a public repository of non-stationary data streams often in use by the community \cite{souza2015data,umer2020comparative}, which contains several stream datasets with incremental drift and differently shaped concepts within their classes. An initial part of each dataset in this repository is considered as supervised ($<5\%$ of the entire stream length). The batch size is fixed by taking into account the total number of streaming instances of the dataset and morphological characteristics of each dataset, so that the number of iterations (batches) will be greater than or equal to $100$. Details of the configurations for every dataset in this repository are given in Table \ref{tab:datasets}, which match the experimental configuration used in related studies \cite{ferreira2019amanda,arostegi2024aigas}. 

Unfortunately, such datasets do not inherently feature abrupt recurring CDs. To address this, we induce such drift events by appending instances at the end of the stream that correspond to a previous distribution of concepts within the stream. Given that these datasets typically exhibit incremental drift, we must ensure that any recurrent drift added at the end of the dataset results in a new concept distribution that can be uniquely distinguished from all previous distributions in the stream, particularly considering the unsupervised nature of stream instances. If the new distribution of instances in the feature space after a recurrent drift can be mapped to two prior distributions of concepts with differing mappings between concepts and classes, this ambiguity may lead to classification errors when the classifier predicts labels for instances in new batches of the stream. To avoid this issue, we have analyzed and identified datasets from the repository where the incremental drift and distribution of instances in the feature space hinder the discriminability of past concepts. The datasets where recurrent drifts cannot be induced for this reason are \texttt{4CRT}, \texttt{4CRE-V1}, \texttt{4CRE-V2}, \texttt{UG3C2D}, \texttt{UG2C3D}, \texttt{UG2C5D}, and \texttt{4CE1CF}.

When it comes to evaluation metrics, we follow the common practice in the area of non-stationary data streams \cite{gama2009issues,gama2013evaluating} and evaluate the algorithms using the so-called \emph{prequential error}. This performance measure is computed as the accumulated sum of a loss function between the predicted and observed values \cite{dawid1984present}, as follows:
\begin{equation} 
P_e(i) = \frac{1}{i} \sum_{k=1}^{i} \mathcal{L}(y_k, \widehat{y_k}) = \frac{1}{i} \sum_{k=1}^{i} e_k, 
\end{equation}
where the prequential error is computed at time $i$, $\mathcal{L}(y_k, \widehat{y_k})$ represents the loss function between the predicted class $\widehat{y_k}$ and the true class $y_k$ for stream instance $k$, and $e_k$ denotes the error for instance $k$. The prequential error enables monitoring the performance evolution of models that adapt over time. Additionally, we report the average F1 score for each dataset, as well as aggregate statistics across all datasets.

To ensure reproducibility, the source code, datasets, and results associated with this paper are available at
\url{https://git.code.tecnalia.com/maria.arostegi/aigas-devl-rc}.

\section{Results and Discussion} \label{sec:results}

The results from our experiments are summarized in Table \ref{tab:Preq} (prequential error), and Table \ref{tab:f1} (macro F1 score). In both tables, the best outcomes for every streaming dataset are shaded in gray. The tables are divided into two subsets of results: one for datasets subject to incremental CD (top) and the other for datasets in which a recurrent CD has been induced (bottom, datasets with \texttt{RCD} suffix). The last two rows of every subset of results inform about the mean and standard deviation statistics of all the methods compared, computed across all datasets within the subset. 
\begin{table}[!h]
	\centering
	\caption{Average prequential error results.}
	\label{tab:Preq}
	\resizebox{0.82\columnwidth}{!}{
	\begin{tabular}{ccccc}
		\toprule
		\textbf{Dataset} & \makecell{\textbf{A-FCP}\\\cite{ferreira2019amanda}} & \makecell{\textbf{A-DCP}\\\cite{ferreira2019amanda}} & \makecell{\textbf{AiGAS-dEVL}\\\cite{arostegi2024aigas}} & \makecell{\textbf{AiGAS-dEVL-RC}\\(proposed)}  \\
		\midrule
        \texttt{1CDT}  & 0.02 & 0.07 & \cellcolor{gray!30}{0.01} & \cellcolor{gray!30}{0.01} \\
\texttt{1CHT}  & \cellcolor{gray!30}{0.33} & 0.38 & 0.36 & 0.36 \\
\texttt{2CDT} & 5.84 & 6.17 & \cellcolor{gray!30}{3.43} & \cellcolor{gray!30}{3.43} \\
\texttt{2CHT}  & 14.38 & 30.48 & \cellcolor{gray!30}{9.85} & \cellcolor{gray!30}{9.85} \\
\texttt{5CVT} & 55.66 & 46.79 & \cellcolor{gray!30}{8.90} & \cellcolor{gray!30}{8.90} \\
\texttt{1CSURR} & \cellcolor{gray!30}{5.10} & 6.12 & 5.20 & 5.20 \\
\texttt{MG2C2D} & 7.89 & 16.59 & \cellcolor{gray!30}{7.50} & \cellcolor{gray!30}{7.50} \\
\texttt{FG2C2D} & 13.91 & 17.71 & \cellcolor{gray!30}{4.40} & \cellcolor{gray!30}{4.40} \\
\texttt{GEARS} & 2.72 & 4.25 & \cellcolor{gray!30}{0.49} & \cellcolor{gray!30}{0.49} \\
\texttt{4CRT} & \cellcolor{gray!30}{0.01} & \cellcolor{gray!30}{0.01} & {0.01} & \cellcolor{gray!30}{0.01} \\
\texttt{4CRE-V1} & 28.28 & 66.82 & \cellcolor{gray!30}{2.45} & \cellcolor{gray!30}{2.45} \\
\texttt{4CRE-V2} & 8.96 & 35.50 & \cellcolor{gray!30}{7.63} & \cellcolor{gray!30}{7.63} \\
\texttt{UG2C2D} & 5.59 & 5.60 & \cellcolor{gray!30}{4.43} & \cellcolor{gray!30}{4.43} \\
\texttt{UG2C3D} & 5.81 & 6.62 & \cellcolor{gray!30}{4.87} & \cellcolor{gray!30}{4.87} \\
\texttt{UG2C5D} & 8.57 & 9.20 & \cellcolor{gray!30}{8.44} & \cellcolor{gray!30}{8.44} \\
\texttt{4CE1CF} & 2.22 & \cellcolor{gray!30}{1.89} & 2.27 & 2.27 \\
\midrule
Average & 10.33 & 15.89 & \cellcolor{gray!30}{4.39} & \cellcolor{gray!30}{4.39} \\
Standard Dev. & 14.01 & 19.41 & \cellcolor{gray!30}{3.33} & \cellcolor{gray!30}{3.33} \\
\midrule
\texttt{1CDT-RCD}  & 3.11 & 5.29 & 0.15 & \cellcolor{gray!30}{0.01} \\
\texttt{1CHT-RCD}  & 5.52 & 5.57 & 0.97 & \cellcolor{gray!30}{0.58} \\
\texttt{2CDT-RCD}  & 15.31 & 35.45 & 13.01 & \cellcolor{gray!30}{3.11} \\
\texttt{2CHT-RCD}  & 22.46 & 45.70 & 18.03 & \cellcolor{gray!30}{9.55} \\
\texttt{5CVT-RCD} & 67.96 & 72.56 & 23.40 & \cellcolor{gray!30}{8.60} \\
\texttt{1CSURR-RCD} & 10.32 & 11.96 & 6.90 & \cellcolor{gray!30}{5.20} \\
\texttt{MG2C2D-RCD} & 17.48 & 23.63 & 16.10 & \cellcolor{gray!30}{7.04} \\
\texttt{FG2C2D-RCD} & 15.47 & 23.77 & 9.70 & \cellcolor{gray!30}{5.72} \\
\texttt{GEARS-RCD} & 2.69 & 4.40 & \cellcolor{gray!30}{0.65} & \cellcolor{gray!30}{0.65} \\
		\midrule
Average & 17.81 & 25.37 & 9.88 & \cellcolor{gray!30}{4.50} \\
Standard Dev. & 20.00 & 22.81 & 8.41 & \cellcolor{gray!30}{3.59} \\
		\bottomrule
	\end{tabular}}
    \vspace{-3mm}
\end{table}

At a first glance, the proposed AiGAS-dEVL-RC approach delivers similar results to AiGAS-dEVL and other methods in the comparison across datasets without recurrent CD. The advantage of AiGAS-dEVL-RC becomes evident when abrupt recurrent drifts occur (datasets ending with \texttt{-RCD}). In such cases, AiGAS-dEVL-RC effectively identifies and retrieves previous concepts, leading to better predictions for new instances and avoiding the catastrophic performance degradation observed in other algorithms. For example, on the dataset \texttt{1CSURR}, A-FCP slightly outperforms AiGAS-dEVL and AiGAS-dEVL-RC in the absence of abrupt recurrent CD (prequential error of 5.10 vs. 5.20). However, when a recurrent CD is introduced (\texttt{1CSURR-RCD}), all benchmarked methods show a significant drop in performance except AiGAS-dEVL-RC, which resiliently adapts by retrieving prior concepts from memory. Remarkably, in several datasets (\texttt{2CDT-RCD}, \texttt{2CHT-RCD}, \texttt{MG2C2D-RCD}, and \texttt{5CVT-RCD}), AiGAS-dEVL-RC achieves better performance than on the corresponding datasets without recurrent CD (\texttt{2CDT}, \texttt{2CHT}, \texttt{MG2C2D}, and \texttt{5CVT}). This improvement is attributed to its memory of stored samples: while the GNG inherently adapts to changes, the retrieval of stored prototypical instances in cases of recurrence further enhances prediction quality. The results with \texttt{GEARS} and \texttt{GEARS-RCD} are noteworthy. Both datasets feature two gears rotating in opposite directions at the same speed, with aligned blades. In \texttt{GEARS}, the rotation direction remains constant, while \texttt{GEARS-RCD} introduces an abrupt CD. AiGAS-RC performs slightly worse on GEARS (0.49 vs. 0.65) because it interprets the CD as a change in motion (rotation direction), rather than as a return to a past distribution.
\begin{table}[!h]
	\centering
	\caption{Average macro F1 results.}
	\label{tab:f1}
	\resizebox{0.85\columnwidth}{!}{
	\begin{tabular}{ccccc}
		\toprule
		\textbf{Dataset} & \makecell{\textbf{A-FCP}\\\cite{ferreira2019amanda}} & \makecell{\textbf{A-DCP}\\\cite{ferreira2019amanda}} & \makecell{\textbf{AiGAS-dEVL}\\\cite{arostegi2024aigas}} & \makecell{\textbf{AiGAS-dEVL-RC}\\(proposed)}  \\
		\midrule
		\texttt{1CDT} &	\cellcolor{gray!30}0.999 & \cellcolor{gray!30}0.999 & \cellcolor{gray!30}0.999 & \cellcolor{gray!30}0.999 \\
\texttt{1CHT}  & \cellcolor{gray!30}0.996 & 0.995 & \cellcolor{gray!30}0.996 & \cellcolor{gray!30}0.996 \\
\texttt{2CDT}  & 0.940 & 0.939 & \cellcolor{gray!30}0.965 & \cellcolor{gray!30}0.965 \\
\texttt{2CHT}  & 0.850 & 0.620 & \cellcolor{gray!30}0.900 & \cellcolor{gray!30}0.900 \\
\texttt{5CVT} & 0.369 & 0.528 & \cellcolor{gray!30}0.916 & \cellcolor{gray!30}0.916 \\
\texttt{1CSURR} & \cellcolor{gray!30}0.946 & 0.935 & 0.941 & 0.941 \\
\texttt{MG2C2D} & 0.918 & 0.820 & \cellcolor{gray!30}0.924 & \cellcolor{gray!30}0.924 \\
\texttt{FG2C2D} & 0.710 & 0.800 & \cellcolor{gray!30}0.942 & \cellcolor{gray!30}0.942 \\
\texttt{GEARS} & 0.970 & 0.950 & \cellcolor{gray!30}0.994 & \cellcolor{gray!30}0.994 \\
\texttt{4CRT} & \cellcolor{gray!30}0.999 & \cellcolor{gray!30}0.999 & \cellcolor{gray!30}0.999 & \cellcolor{gray!30}0.999 \\
\texttt{4CRE-V1} & 0.717 & 0.331 & \cellcolor{gray!30}0.975 & \cellcolor{gray!30}0.975 \\
\texttt{4CRE-V2} & 0.910 & 0.644 & \cellcolor{gray!30}0.923 & \cellcolor{gray!30}0.923 \\
\texttt{UG2C2D} & 0.944 & 0.944 & \cellcolor{gray!30}0.955 & \cellcolor{gray!30}0.955 \\
\texttt{UG2C3D} & 0.943 & 0.936 & \cellcolor{gray!30}0.951 & \cellcolor{gray!30}0.951 \\
\texttt{UG2C5D} & 0.914 & 0.907 & \cellcolor{gray!30}0.919 & \cellcolor{gray!30}0.919 \\
\texttt{4CE1CF} & 0.975 & \cellcolor{gray!30}0.980 & 0.977 & 0.977 \\
		\midrule
        Average	& 0.881 & 0.833 & \cellcolor{gray!30}0.955 & \cellcolor{gray!30}0.955 \\
Standard Dev. & 0.162 & 0.199 & \cellcolor{gray!30}0.033 & \cellcolor{gray!30}0.033 \\
\midrule
\texttt{1CDT-RCD}  & 0.960 & 0.930 & 0.998 & \cellcolor{gray!30}0.999 \\
\texttt{1CHT-RCD}  & 0.929 & 0.928 & 0.990 & \cellcolor{gray!30}0.994 \\
\texttt{2CDT-RCD}  & 0.816 & 0.540 & 0.834 & \cellcolor{gray!30}0.968 \\
\texttt{2CHT-RCD}  & 0.740 & 0.430 & 0.784 & \cellcolor{gray!30}0.904 \\
\texttt{5CVT-RCD} & 0.223 & 0.259 & 0.760 & \cellcolor{gray!30}0.920 \\
\texttt{1CSURR-RCD} & 0.890 & 0.874 & 0.909 & \cellcolor{gray!30}0.942 \\
\texttt{MG2C2D-RCD} & 0.826 & 0.750 & 0.837 & \cellcolor{gray!30}0.929 \\
\texttt{FG2C2D-RCD} & 0.660 & 0.740 & 0.879 & \cellcolor{gray!30}0.925 \\
\texttt{GEARS-RCD} & 0.970 & 0.950 & \cellcolor{gray!30}0.993 & \cellcolor{gray!30}0.993 \\
        \midrule
Average	& 0.779 & 0.711 & 0.887 & \cellcolor{gray!30}0.953 \\
Standard Dev. & 0.233 & 0.248 & 0.091 & \cellcolor{gray!30}0.036 \\
		\bottomrule
	\end{tabular}}
    \vspace{-1mm}
\end{table}

\paragraph*{Limitations} The main limitations arise from AiGAS-dEVL \cite{arostegi2024aigas}, particularly due to the Growing Neural Gas (GNG) algorithm. GNG can become computationally intensive with large or high-dimensional datasets, especially when cluster structures are irregular or overlapping. Its worst-case quadratic complexity with respect to the number of nodes limits scalability and increases resource demands, especially when adapting to continuous data streams. On the other hand, some of the algorithmic choices made in AiGAS-dEVL-RC are suited to deal with tabular datasets, from the clustering approach producing the prototypical instances, the distance function $D(\cdot,\cdot)$ used in the {\color{darkspringgreen}\textbf{\texttt{retrieve}}} and {\color{denim}\textbf{\texttt{store}}} phases, or the classifier used to annotate the GNG node labels and the batch instances themselves. However, the compounding phases of AiGAS-dEVL-RC should be regarded as a methodological workflow in which such choices are configured depending on the characteristics of the dataset at hand (e.g. drift speed, severity, dimensionality, or semantic meaning of stream instances, among others). Automating the configuration process of the different algorithms involved in each phase will be part of the research directions to be tackled in the future, jointly with means to balance the trade-off between the representability of distributions stored in the memory and its memory footprint.

\section{Conclusions and Future Work} \label{sec:conclusions}

This work has explored the use of unlabeled information in non-stationary data streams under two specific circumstances: extreme verification latency and both incremental and abrupt recurring concept drifts. Predicting data streams under these circumstances is challenging because the model must balance adaptability to changing data distributions with the retention of relevant past knowledge while working under real-time constraints. Additionally, the absence of immediate labels complicates model updates, drift detection, and evaluation. 

To advance over these challenges, we have introduced a novel approach (AiGAS-dEVL-RC) which leverages the proven adaptability of GNGs to learn from the stream in an online fashion and to store the information necessary to identify previous concept distribution. AiGAS-dEVL-RC reuses the stored knowledge in the presence of recurring abrupt CDs. Additionally, two key elements have been emphasized: (1) only knowledge (in the form of neural gas nodes, their predicted labels, and centroid-related information) is stored when it differs significantly from previously stored concept distributions, preventing memory issues; and (2) the identification of previous knowledge that overlaps with the prevalent concept distribution in the stream relies on the similarity of their $\alpha$-shapes based on the Intersection over Union metric. Our experiments have considered methods from the literature on drifting data streams under EVL. The results reveal that AiGAS-dEVL-RC not only achieves predictive performance comparable to its predecessor (AiGAS-dEVL \cite{arostegi2024aigas}) but also demonstrates greater resilience to abrupt, recurrent drifts. 

Future research will focus on two main directions: (1) developing methods to automatically detect and resolve ambiguities, potentially through active supervision and the integration of additional data into the model’s knowledge base; and (2) extending AiGAS-dEVL-RC to handle more complex data modalities, such as video, text, and multivariate time series. By incorporating mechanisms to integrate and process heterogeneous data types (e.g., textual, visual, and sensor data) into the compounding steps of AiGAS-dEVL-RC (namely, {\color{darkgoldenrod}\textbf{characterize}}, {\color{darkspringgreen}\textbf{retrieve}}, {\color{darkred}\textbf{predict}}, and {\color{denim}\textbf{store}}), and by automating the configuration of the algorithms involved in each step, AiGAS-dEVL-RC can be enhanced to tackle real-world scenarios involving diverse concept drifts and non-tabular data flows.

\section*{Acknowledgments}

The authors would like to thank the European Commission under the European Defence Fund (EDF-2021-DIGIT-R) for its funding support through the FaRADAI project (ref. 101103386), and the Basque Government through BEREZ-IA (KK-2023/00012) and MATHMODE (IT1456-22). During the writing process, the authors used large language models to improve the readability of the paper. After using them, the authors reviewed and edited the content as needed, assuming full responsibility for the content of the published article.

\bibliographystyle{IEEEtran}
\bibliography{main.bib}
\vfill
\end{document}